\documentclass[]{spie}  %>>> use for US letter paper
\usepackage{amsmath,amsfonts,amssymb}
\usepackage{graphicx}
\usepackage[colorlinks=true, allcolors=blue]{hyperref}
\usepackage{graphicx}
\usepackage{amsfonts}       % blackboard math symbols
\usepackage{color}
\usepackage{amsmath}
\usepackage{subfig}
\usepackage{graphicx}

\newcommand{\vect}[1]{\boldsymbol{#1}}
\title{Projection Image-to-Image Translation in Hybrid X-ray/MR~Imaging}

\author[a,b]{Bernhard~Stimpel}
\author[a,b]{Christopher~Syben}
\author[a]{Tobias~W\"urfl}
\author[a]{Katharina~Breininger}
\author[a,b]{Jonathan~M.~Lommen}
\author[b]{Arnd~D\"orfler}
\author[a]{Andreas~Maier}
\affil[a]{Pattern~Recognition~Lab, Friedrich-Alexander-Universit\"at Erlangen-N\"urnberg, Germany}
\affil[b]{Department~of~Neuroradiology, Friedrich-Alexander-Universit\"at Erlangen-N\"urnberg, Germany}

\authorinfo{Further author information:\\Bernhard~Stimpel: bernhard.stimpel@fau.de}

\begin{document} 
\maketitle

\begin{abstract}
	The potential benefit of hybrid X-ray and MR imaging in the interventional environment is large due to the combination of fast imaging with high contrast variety. However, a vast amount of existing image enhancement methods requires the image information of both modalities to be present in the same domain. To unlock this potential, we present a solution to image-to-image translation from MR projections to corresponding X-ray projection images. The approach is based on a state-of-the-art image generator network that is modified to fit the specific application. Furthermore, we propose the inclusion of a gradient map in the loss function to allow the network to emphasize high-frequency details in image generation. Our approach is capable of creating X-ray projection images with natural appearance. Additionally, our extensions show clear improvement compared to the baseline method.  
\end{abstract}

\section{Introduction}
Hybrid imaging exhibits high potential in diagnostic and interventional applications \cite{Fahrig2001}. Future advances in research may leverage the combination of Computed Tomography (CT) and Magnetic Resonance Imaging (MRI) to clinical applicability. Especially for interventional purposes, the gain from simultaneously acquiring soft- and dense-tissue information would yield great opportunities. Assuming the information of both modalities is present at the same time, numerous existing post-processing methods would become applicable. Image fusion techniques, e.g., image overlays, have proven useful in the past. Additionally, one can think about image enhancement techniques, e.g., image de-noising or super-resolution. 
To enable the latter methods, it is beneficial to have the data available in the same domain. Solutions to generate CT images from corresponding MRI data were presented previously \cite{Navalpakkam2013,Nie2017}, mostly in order to generate attenuation maps for radiation therapy. However, all of these are applied to volumetric data, i.e., slice images. In contrast, interventional procedures rely heavily on line integral data from X-ray projection imaging. Projection images which exhibit the same perspective distortion can also be acquired directly using an MR device \cite{Syben2017}. This avoids time-consuming volumetric acquisition and subsequent forward projection. The synthesis of the desired X-ray projections from the corresponding MRI signal is an inherently ill-posed problem. Large portions of the dominant signal in X-ray are obtained from bone which provides little to no signal in MRI. Furthermore, because air also provides no signal in MRI the intensity ranges of both materials overlap which accounts for an even more complicated differentiation. The information for the generation of accurate intensity values can, therefore, solely be drawn from the structural information that is present in the image. In case of volumetric imaging the materials may be unknown prior to the synthesis but they are resolved in distinct regions in the image. In contrast, in projection imaging this structural information diminishes by integration of the intensity or attenuation values on the detector. This corresponds to a linear combination of multiple slice images with unknown path length which further increases the difficulty of the synthesis task. To the best of our knowledge, no solution to this problem was proposed up to now. Therefore, we investigate a solution to generate X-ray projections from corresponding MRI views through image-to-image translation.

\section{Methods}
\label{sec:methods}
Current methods for the synthesis of volumetric CT data from corresponding MR scans are typically atlas- or learning based.
%CT data synthesis from corresponding MR acquisitions is currently performed based on volumetric data for treatment planning in radiotherapy. Most approaches to this task are either atlas- or learning-based methods. 
Atlas-based methods, however, are difficult to apply to projection imaging due to the high variance in appearance resulting from differences in the projection geometry. In contrast, learning-based solutions exploit image-level features only. Considering the aforementioned difficulties regarding the synthesis of X-ray projections from MR signal, the use of generative adversarial networks (GANs) is suitable. The adversarial design allows for a more extensive exploration of the possible solution space. Consequently, we propose a deep learning solution using a generative adversarial network for this image-to-image-translation task. 

\subsection{Network architecture} 
\begin{figure}
	\centering
	\includegraphics[width=0.8\textwidth]{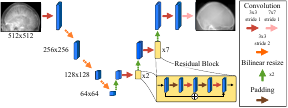}	
	\caption{The proposed architecture for the generator network.}
	\label{fig:architecture}
	\vspace{-10pt}
\end{figure}

A feed-forward network is used as the image generator. Our network's design is based on the architecture proposed by Johnson et al. \cite{Johnson2016}, which was also adopted for various other applications, e.g., \cite{Wanga}. The architecture is designed in an encoder-decoder fashion and is fully convolutional. In the original manuscript, multiple residual blocks are introduced at the lowest resolution level of the network. This results in the accumulation of a large portion of the available model capacity at these coarse resolution layers that determine the general structures and their arrangement in the image. Due to the vast amount of possibilities in natural imaging synthesis this may be intuitive. However, the underlying variance in medical projection images is only small compared to natural image scenes.  %However, the underlying variance in medical projection images is only small compared to natural image scenes. 
Additionally, during interventional treatments, valuable information is largely drawn from high-frequency details such as contrast and clear edges. These indicate the outline of bones, medical devices, and similar structures. To shift the capabilities of the generator network in favor of these high-frequency structures, we distribute the residual blocks across higher resolution levels instead. 
% To this end, we distribute the residual blocks at higher resolution levels instead. 
% Aggregating a large portion of the model capacity at the most coarse resolution is, therefore, unintuitive. Instead, we distribute the residual blocks at higher resolution levels.
Furthermore, bilinear upscaling is used in place of the transposed convolution operation, which was recently related to checkerboard artifacts~\cite{Odena2016}. A visualization of the final architecture is shown in Fig.~\ref{fig:architecture}.
\subsection{Objective function}
The main part of the proposed objective function is based on the discriminator network which is trained to separate the generated fake images from the real label images. For the discriminator we stick with the proven architecture proposed in \cite{Zhu2017}. Although an adversarial loss is powerful, it is also less constrained to the target image than conventional cost functions. Resulting from this, it is often used in combination with a second metric. The generator network is designed such that the focus is on generating accurate high-frequency structures. Consequently, the objective function for the optimization process must be chosen accordingly. Pixel-wise metrics, e.g., L1- or L2-norm, which are frequently used for attenuation map generation in radiotherapy, do not satisfy this requirement and are related to blurrier results in comparison \cite{Dosovitskiy2016}. Considering the importance of high-frequency structures, using a feature matching loss as proposed by \cite{Johnson2016} is suitable. This loss functions is based on the extraction and comparison of high-level image features between the generated and the reference image. 
In recently published work, \cite{Stimpel2017a} concluded that utilizing the VGG-19 network pre-trained on ImageNet for the computation of this feature matching loss is appropriate for medical projection images. 
The aforementioned high-level image features that are used to compute the feature matching loss exhibit many edges and similar structures. However, the majority of the projection images consist of homogeneous regions. Additional emphasis of high-frequency details is achieved by including an edge-weighting to the loss computation. First, a gradient map of the label image is computed using the Sobel filter \cite{Sobel1968}. Second, this gradient map is used to weight the loss such that the loss generated from edges is emphasized and that from homogeneous regions is attenuated. Starting with the GAN-loss, this can be formulated mathematically as

%To emphasize the influence of high-frequency details, we include a gradient map of the label images into the optimization process. Subsequently, this map is used to weight the loss such that the loss generated from edges is emphasized and that from homogeneous regions is attenuated. Mathematically, this can be formulated as
\begin{equation}
\label{eq:gan_loss}
\ell_\text{GAN}(\vect{L},\vect{G}, D) = \mathbb{E}_{\vect{L},\vect{G}}\left[	\log D(\vect{L},\vect{G})\right] +  \mathbb{E}_{\vect{L},\vect{G}}\left[1-	\log D(\vect{G})\right]
\end{equation}
where $D$ is the discriminator network and  $\vect{L}$ and  $\vect{G}$ are the label and generated image, respectively. The second part of the loss function is the feature matching loss described by
\begin{equation}
\label{eq:fm_loss}
\ell_\text{FM}(\vect{L},\vect{G}) = \sum_{s}^{S} \left( \vect{V}_s(\vect{L}) - \vect{V}_s(\vect{G}) \right) \;,
\end{equation}
where $\vect{V}_s(\vect{L})$ and $\vect{V}_s(\vect{G})$ are the feature activation maps of the VGG-19 network at the layer $s \in S$. This leads to the final objective function
\begin{equation}
\label{eq:edge_loss}
\ell(\vect{L},\vect{G},D) = \left(\ell_\text{GAN}(\vect{L},\vect{G}, D) + \ell_\text{FM}(\vect{L},\vect{G})  \right) \cdot \vect{E_L}\;,
\end{equation}
which is the combination of the GAN and feature matching loss weighted by the gradient map $\vect{E_L}$ of the label image.
%\red{ -  Scale $ \vect{E_L}_s$ not yet correct}% computed using the Sobel filter. 
\subsection{Data and Experiments}
Both, MRI and CT scans of four individuals with different pathologies were provided % by the Department of Neuroradiology, University Hospital Erlangen 
(MR: 1.5\,T MAGNETOM Aera / CT: SOMATON Definition, Siemens Healthineers, Erlangen / Forchheim, Germany). 
The tomographic data is registered using 3D Slicer and forward projections are generated using the CONRAD framework~\cite{Maier2013}. All projections are zero-centered and normalized prior to training. For the input data, i.e., the MRI projections, this preprocessing is applied on each subject individually and not on the whole dataset to account for differences in the MR protocols. Training was performed for a fixed number of 400 epochs using the ADAM optimizer.
Evaluation of the proposed approach is performed quantitatively as well as qualitative. Because of the limited data available, a 4-fold cross validation is computed, i.e., for each evaluation the projections of three patient data sets are used for training and one for testing. To this end, projections from a $180^{\circ}$ rotation in the transversal plane are created with a projection geometry that closely resembles common clinical X-ray systems. For evaluation, the mean squared error (MSE), structural similarity index (SSIM), and peak signal-to-noise ratio (PSNR) are calculated. Furthermore, we investigate how the performance with respect to these metrics depends on the projection angle. All projections are normalized beforehand. For MSE and PSNR only pixel are considered that are nonzero in the label images to limit the optimistic bias caused by the large homogeneous air regions. We also compare our approach to the originally proposed architecture as used for example in \cite{Johnson2016,Wanga}, which we will refer to as "baseline".   
\section{Results}
\label{sec:results}
\begin{figure}
	\centering	
	\subfloat[MSE over different projection angles in degrees.]{\includegraphics[width=0.47\textwidth]{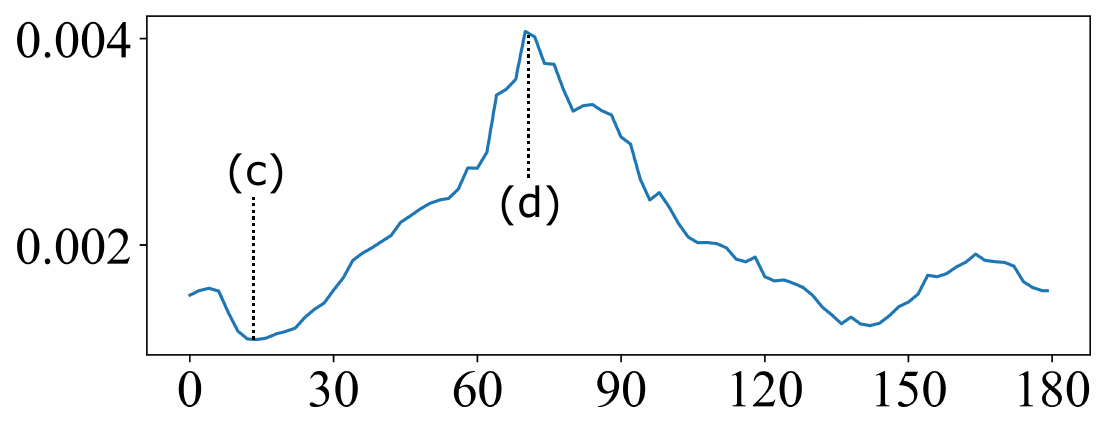}%
		\label{fig:angular_mse}}
	\hfil
	\subfloat[SSIM over different projection angles in degrees.]{\includegraphics[width=0.47\textwidth]{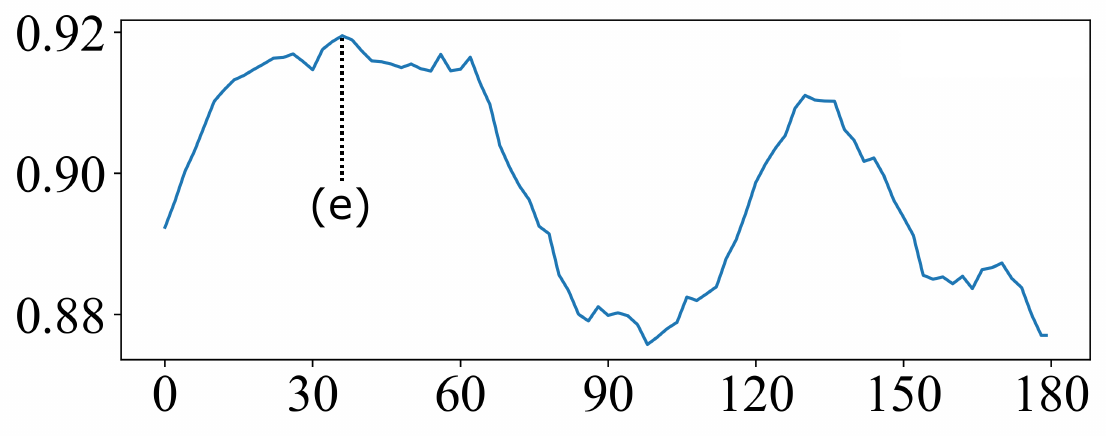}%
		\label{fig:angular_ssim}}	
	\\
	\subfloat[]{\includegraphics[width=0.3\textwidth]{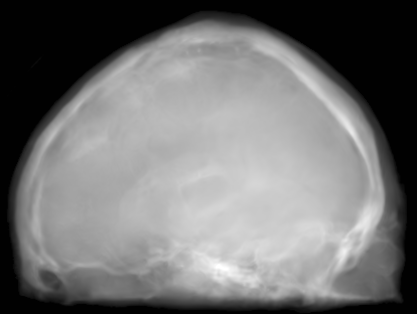}%
		\label{fig:mse_8}}
	\hfil
	\subfloat[]{\includegraphics[width=0.3\textwidth]{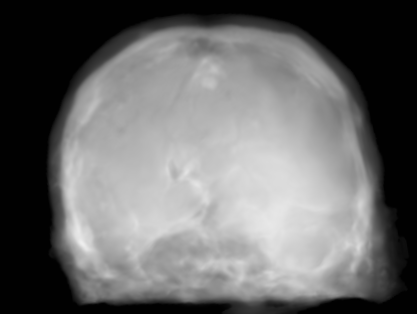}%
		\label{fig:mse_36}}
	\hfil
	\subfloat[]{\includegraphics[width=0.3\textwidth]{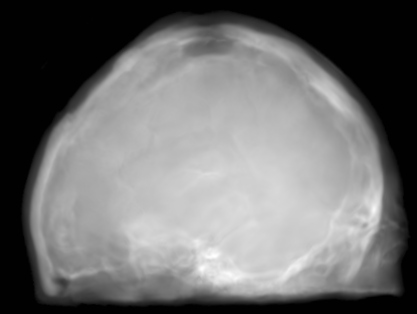}%
		\label{fig:ssim_18}}	
	\\
	\caption{Evaluation metrics of the projections along a circular trajectory (a-b). Example projections at selected projection angles (c-e).}
	\vspace{-10pt}
\end{figure}

\begin{figure}
	\centering	
	\text{MRI projection}%	
	\hfil
	\text{Generated projection}%
	\hfil
	\text{Label projection}%
	\hfil
	\\
	\subfloat[Patient 1 at 15$^\circ$ projection angle]{
		\includegraphics[width=0.295\textwidth]{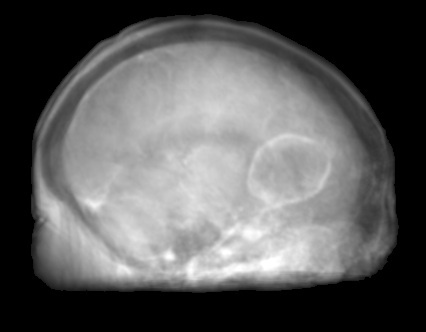} \hfil%
		\includegraphics[width=0.295\textwidth]{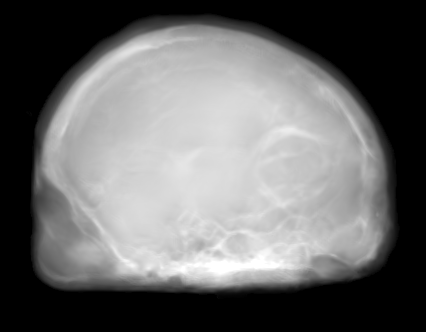}  \hfil% \\
		\includegraphics[width=0.295\textwidth]{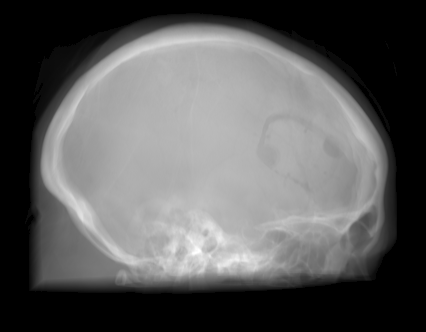} %		
		\label{fig:Pat6}	
	}
	\\
	\subfloat[Patient 3 at 95$^\circ$ projection angle]{
		\includegraphics[width=0.295\textwidth]{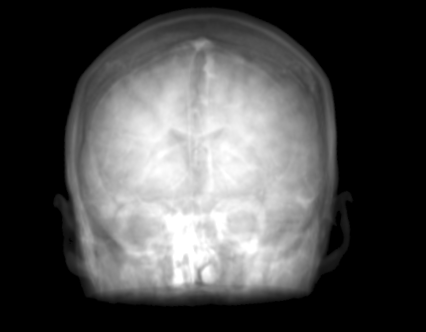} \hfil%
		\includegraphics[width=0.295\textwidth]{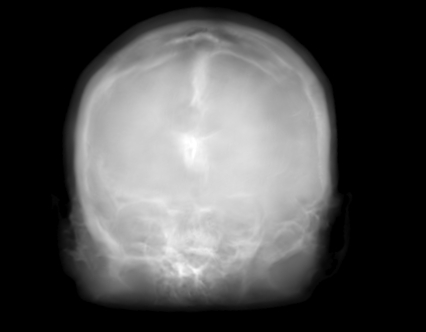}  \hfil% \\
		\includegraphics[width=0.295\textwidth]{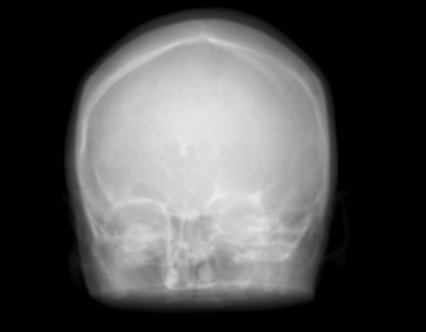} %		
		\label{fig:Pat8}	
	}
	\\
	\subfloat[Patient 4 at 360$^\circ$ projection angle]{
		\includegraphics[width=0.295\textwidth]{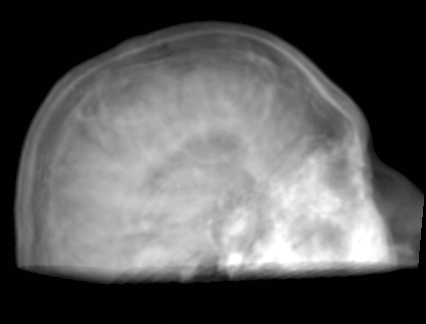} \hfil%
		\includegraphics[width=0.295\textwidth]{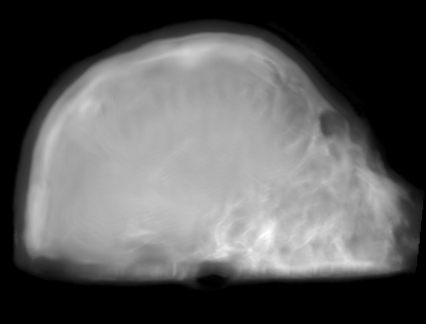}  \hfil% \\
		\includegraphics[width=0.295\textwidth]{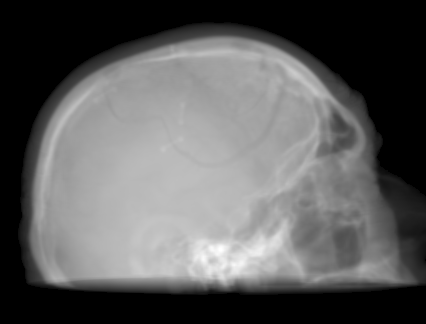} %		
		\label{fig:Pat10}
	}
	\caption{Representative examples of the projection image-to-image translation for different projection angles and patients.}
	\label{fig:projection_results}
\end{figure}

\begin{figure}
	\centering
	\subfloat[Baseline]{\includegraphics[width=0.295\textwidth]{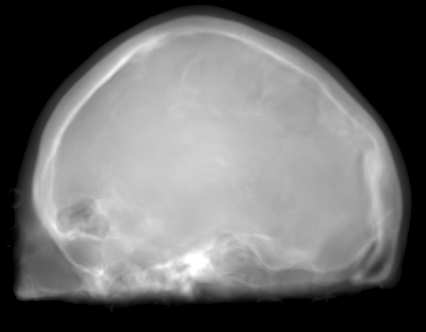}%
		\label{fig:7_resnet}}
	\hspace{1pt}
	\subfloat[Our edge-weighted]{\includegraphics[width=0.295\textwidth]{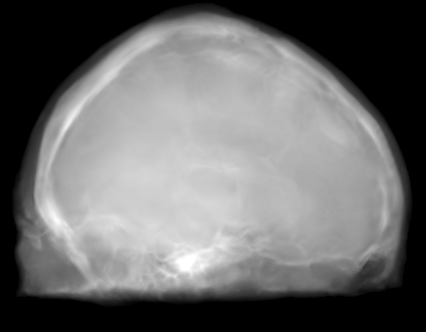}%
		\label{fig:7_mnet}}
	\hspace{1pt}
	\subfloat[Label X-ray]{\includegraphics[width=0.295\textwidth]{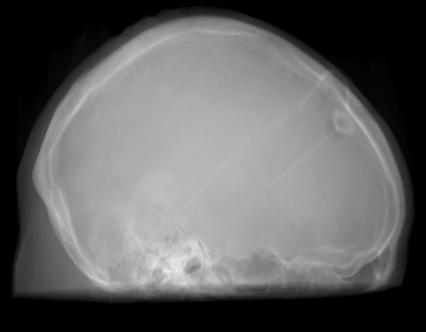}%
		\label{fig:7_label}}
	\hspace{1pt}	
	\caption{Comparison of the projections generated by the different network architectures based on Patient 2.}
	\label{fig:networks_results}	
\end{figure}

The proposed approach was successful in generating X-ray projections with a contrast similar to the one seen in true fluoroscopic X-ray images. Quantitative results of the generated projection images for all patients are presented in Tab.~\ref{tab:eval_patients} and for the different network architectures in Tab. \ref{tab:eval_networks}. In Fig. \ref{fig:angular_mse} to \ref{fig:ssim_18} the behavior of the MSE and SSIM w.r.t the projection angle is presented. Additional qualitative results of the proposed projection image-to-image translation pipeline for different patient data sets are shown in Fig.~\ref{fig:Pat6} to \ref{fig:Pat10}. In Fig.~\ref{fig:networks_results} the influence of the modified network architecture, as well as the weighted loss w.r.t. to the edge map are presented. 
\begin{table}[h]
	\centering
	\caption{MSE, SSIM, and PSNR of our edge-weighted approach for all datasets.}
	\begin{tabular*}{0.9\linewidth}{l@{\extracolsep{1cm}}cccc}
		\hline
		&   	MSE  & 				SSIM & 				PSNR  \\ \hline
		Patient 1 &		$0.007\,\pm\,0.001$		& 	$0.894\,\pm\,0.026$	& 	$21.61\,\pm\,0.93$	&   \\ \hline
		Patient 2 &		$0.006\,\pm\,0.003$ 	& 	$0.898\,\pm\,0.015$	&	$22.49\,\pm\,2.03$ 	&	\\ \hline
		Patient 3 &		$0.010\,\pm\,0.002$ 	&	$0.892\,\pm\,0.013$	&	$20.31\,\pm\,0.99$	&	\\ \hline
		Patient 4 &		$0.017\,\pm\,0.004$		&	$0.872\,\pm\,0.014$	&	$17.79\,\pm\,1.08$	&	\\ \hline		
	\end{tabular*}
	\label{tab:eval_patients}	
	\centering
	\vspace{5pt}
	\caption{MSE, SSIM, and PSNR of the different network architectures.}
	\begin{tabular*}{0.9\linewidth}{l@{\extracolsep{0.2cm}}rrrr}
		\hline
		\multicolumn{1}{l}{} & 
		\multicolumn{1}{c}{MSE} & 
		\multicolumn{1}{c}{SSIM} & 
		\multicolumn{1}{c}{PSNR} & 
		\\ \hline
		%&   	MSE  & 				SSIM & 				PSNR  \\ \hline
		Reference Architecture				&	$0.010\,\pm\,0.004$			& 	$0.889\,\pm\,0.015$				& 	$20.22\,\pm\,1.78$	&   \\ \hline
		Ours w/o edge-weighting	&	$0.009\,\pm\,0.002$ 		& 	$0.884\,\pm\,0.011$				&	$20.50\,\pm\,1.14$ 	&	\\ \hline
		Ours w/ edge-weighting	&	$\mathbf{0.006\,\pm\,0.003}$& 	$\mathbf{0.898\,\pm\,0.015}$	&	$\mathbf{22.49\,\pm\,2.03}$ 	&	\\ \hline	
	\end{tabular*}
	\label{tab:eval_networks}	
\end{table}
\section{Discussion}
\label{sec:discussion}
The improvement in our method compared to the baseline method is demonstrated by a decreased MSE, increased SSIM, and PSNR in Tab. \ref{tab:eval_networks}. When examining Fig. \ref{fig:7_resnet} to \ref{fig:7_label}, improvements can be observed in the overall increased contrast of high-frequency details. Using the originally proposed architecture \mbox{\cite{Johnson2016,Wanga}}, which gathers the residual blocks at the lowest resolution level, results in overall blurrier results and missing bone structures as seen in Fig.~\ref{fig:7_resnet}. In contrast, the projections generated with the edge-weighted loss resemble the label images more closely. This can especially be observed at the base of the head. The projections created without the weighting also produce many high-frequency details in this region, however, these are less specific in comparison with the edge-weighted results. This results in decreased MSE and increased SSIM and PSNR of the projections synthesized using our approach. In addition, unnatural holes in the brain are generated by the baseline architecture. 
A possible explanation for the fluctuations in the error measure shown in Fig. \ref{fig:angular_mse} and \ref{fig:angular_ssim} is that in our trajectory in the angles around 45 and 135 degrees the projection rays are cast from the side through the brain while around 90 and 180 degrees the angle of incidence is from the front or back side of the skull. In the first case this results in projections that exhibit large homogeneous areas which are easier to synthesize. In the second case, however, the high-frequency edges from the eyes, jawbone, etc. are the dominant structures in the image.
A limiting factor of this study is the low number of patient datasets available. However, the amount of variation introduced by forward projecting the volumes is large. Varying projective geometries account for distinctively different structural appearance of the resulting projections. 
What is of course not covered by these transformations are unique characteristics of individual patients or different pathologies. To investigate the possible translation outcome of these properties larger datasets are required in the future. Also details that are not visible in the MRI projections can not be transferred to the generated images. An example would be interventional devices that are X-ray but not MR sensitive. Regarding subsequent post-processing applications, the question arises how this missing information in the generated projection images should be dealt with, which is subject to future work.  

\section{Conclusion}
\label{sec:conclusion}
We presented an approach to synthesize X-ray projection images from corresponding MRI projections. The proposed redistribution of model capacity at higher resolution layers and the weighting of the computed loss by a gradient map show clear improvements over the baseline method derived from natural image synthesis.
%The proposed extensions of the image-to-image translation pipeline with regards to the baseline method derived from natural image synthesis showed improvements in the generated output.
Increasing the dataset size in subsequent work could help to translate also patient specific details, e.g., pathologies, between the domains. With future advances in hybrid X-ray and MR imaging, this domain transfer can be used to apply valuable post-processing methods. 

\paragraph{Acknowledgment:}
This work has been supported by the project P3-Stroke, an EIT Health innovation project. EIT Health is supported by EIT, a body of the European Union. Additional financial support for this project was granted by the Emerging Fields Initiative (EFI) of the Friedrich-Alexander University Erlangen-N\"urnberg (FAU). Furthermore, we thank the NVIDIA Corporation for their hardware donation. 

\bibliography{C:/Users/stimpel/Documents/Literature_short/library}
\bibliographystyle{spiebib} % makes bibtex use spiebib.bst

\end{document}